\documentclass[journal]{IEEEtran}
\ifCLASSINFOpdf
\else
\fi
\usepackage{makecell}
\usepackage{cite}
\usepackage{graphicx}
\usepackage{multirow}
\usepackage{algorithmic}
\usepackage[ruled]{algorithm2e}
\newsavebox{\tablebox}
\usepackage{lineno,hyperref}
\usepackage{algorithmic}
\usepackage[ruled]{algorithm2e}
\usepackage{setspace}
\usepackage{amsmath,amssymb,amsfonts}
\hypersetup{hidelinks} 
\usepackage{color}
\usepackage{array}

\begin{document}
%
\title{Multi-objective Feature Selection with Missing Data in Classification}
%
%
%

\author{Yu Xue,~\IEEEmembership{Member,~IEEE,}
	Yihang Tang,
	Xin Xu,
	Jiayu Liang,~\IEEEmembership{Member,~IEEE,}
	Ferrante Neri,~\IEEEmembership{Senior Member,~IEEE}

\thanks{Yu Xue, Yihang Tang and Xin Xu are with the School of Computer and Software, Nanjing University of Information Science and Technology, Nanjing 210044, China (e-mail: xueyu@nuist.edu.cn; tangyh@nuist.edu.cn; xuxin1975715@hotmail.com)}
\thanks{Jiayu Liang; Tianjin Key Laboratory of Autonomous Intelligent Technology and System, Tiangong University, Tianjin 300387, China (e-mail:yyliang2012@hotmail.com) }
\thanks{Ferrante Neri; School of Computer Science, University of  Nottingham, UK (email: ferrante.neri@nottingham.ac.uk)}
}

\maketitle

\begin{abstract}

	Feature selection (FS) is an important research topic in machine learning. Usually, FS is modelled as a+ bi-objective optimization problem whose objectives are: 1) classification accuracy; 2) number of features. 
	One of the main issues in real-world applications is missing data. Databases with missing data are likely to be unreliable. Thus, FS performed on a data set missing some data is also unreliable. In order to directly control this issue plaguing the field, we propose in this study a novel modelling of FS: we include reliability as the third objective of the problem. In order to address the modified problem, we propose the application of the  non-dominated sorting genetic algorithm-III (NSGA-III). We selected six incomplete data sets from the University of California Irvine (UCI) machine learning repository. We used the mean  imputation method to deal with the missing data. In the experiments, k-nearest neighbors (K-NN) is used as the classifier to evaluate the feature subsets. Experimental results show that the proposed three-objective model coupled with NSGA-III efficiently addresses the FS problem for the six data sets included in this study.
\end{abstract}


%
\IEEEpeerreviewmaketitle

\section{Introduction}
%
%
%
%
\IEEEPARstart{A}{}\textcolor{black}{large number of data sets contains a lot of irrelevant or redundant features (useless features). Useless features not only waste computing cost, but also decrease the performance of classification} \cite{gheyas2010feature}.  Without prior information about the data, useless features are often difficult to be identified. Useless features are detrimental during classification tasks since they often lead to a low  accuracy and high computational cost. Feature selection (FS) is the process of identification and elimination of these useless features. 

In the past few decades, researchers have proposed many heuristic FS methods. With respect to the logic used to assess the quality of the selected feature (feature subset), they are categorised as 1) filter \cite{hancer2018differential} and 2) wrapper \cite{huang2007hybrid} methods. Filter methods use some specific functions to evaluate the usage of the features. According to the different evaluation functions, the filter methods can be divided into distance-based, consistency-based, dependency-based and information-based. Some well-known filter methods belonging to this category are Koller's \cite{ilprints208}, Relief\cite{subbotin2018quasi} and Set Cover\cite{dash1997feature}. Unlike filter methods, wrapper methods \cite{mafarja2018whale} make use of a specific classifier as its evaluation function, and use classification accuracy to evaluate the candidate selected features. Wrapper methods are usually more accurate than filter methods, since they directly take the classifier as the evaluation function for feature subsets. On the other hand, the employment of the classifier is time-consuming.

In order to be effectively applied, wrapper methods are usually coupled with meta-heuristics that search the space by trying to perform the least function call the possible. Meta-heuristics used to address FS problems can be  divided into two categories 1) those methods that encode FS as a single-objective problem by identifying a specific feature or through the linear combination of multiple objectives; 2) those methods that attempt to simultaneously address multiple criteria of FS and make use of multi-objective optimization algorithms to solve it.

Some examples of meta-heuristics that treat FS as a single-objective problem are brain storm optimization  \cite{sun2013optimal}, differential evolution  \cite{xue2015survey}, artificial bee colony (ABC) \cite{hancer2018pareto} and particle swarm optimization (PSO) \cite{du2019network}. Some other studies, while still modelling FS as a single-objective problem, embed accuracy and solution size within the algorithmic logic.  For example, Zhang $et$ $al$. \cite{tran2018variable} presented a variable length PSO to make the particles have different shorter lengths. Xue $et$ $al$. \cite{xue2020self} devised a PSO algorithm with adaptive parameters and strategies for FS with multiple classifiers. Besides, Xue $et$ $al$. \cite{xue2019self1} also proposed a self-adaptive PSO for FS with large-scale data sets, in order to strengthen the ability of PSO in solving FS problem. In addition, to prevent the loss of excellent offspring, Zhang $et$ $al$. \cite{zhang2015feature} designed a new memory strategy and applied it to bare bones PSO so as to balance the exploration ability of the algorithm.

In fact, FS can be seemed as a multi-objective optimization problem  \cite{tian2020solving,tian2019evolutionary}. For example, sometimes the classification accuracy is high but the number of features is also large, so multiple objectives need to be considered at the same time.  For FS problems, the multi-objective FS methods can provide sets of relatively optimal solutions instead of a single solution. Some popular multi-objective algorithms \cite{li2018two,habib2019multiple,li2019variable,cheng2016reference,lin2018clustering} such as non-dominated sorting genetic algorithm II (NSGA-II) \cite{deb2002fast}, multi-objective evolutionary algorithm with domain decomposition (MOEA/D) \cite{zhang2007moea} and multi-objective PSO \cite{zhang2018competitive ,hu2020multiobjective} are often used for the multi-objective FS problems.  

Recently, the problem of FS has been addressed by coding it as a multi-objective problem. For example, Xue $et$ $al$. \cite{xue2012particle} studied two ideas of integrating non-dominated sorting strategy and crowding strategy into PSO respectively. Besides, for the multi-objective FS problem, Nguyen $et$ $al$. \cite{nguyen2016new} proposed a hybrid/memetic PSO algorithm whose search potential has been augmented by a local search.  Zhang  $et$ $al$. \cite{nguyen2019multiple} observed that FS intrinsically contains multiple conflicting objectives and thus proposed an improved version of MOEA/D. Hancer $et$ $al$. \cite{hancer2018pareto} proposed a multi-objective version of ABC for FS. To solve the problem of multi-objective FS efficiently, Zhang $et$ $al$. \cite{wang2020multi} improved ABC with a parameterless search mechanism and designed a new multi-objective FS algorithm. Moreover, Zhang $et$ $al$. \cite{zhang2015multi} proposed a multi-objective PSO based on the cost of features, which takes the time cost and classification accuracy as two objectives. All the above studies only considered two objectives. In most cases, the FS objectives considered simultaneously are classification accuracy and solution size.

\textcolor{black}{Many studies about FS considered complete data sets.} However, in the real-world applications, \textcolor{black}{missing data} is a common phenomenon due to various unexpected reasons. For example, in the investigation and study, the data may be missing due to negligence of the researcher, the cost of obtaining data is too high, personal privacy involved in data collection, and so on. Missing data have an impact on the formulation of FS problems as it may select unreliable features. To solve the FS problems with missing data, we need to firstly deal with the incomplete data sets. A popular approach to deal with missing data is data imputation, i.e. an interpolation approach that reconstructs missing data on the basis present in the data set. The common imputation methods include mean  imputation method \cite{plaia2006single}, regression imputation method \cite{allison2000multiple}, hot deck imputation method \cite{andridge2010review}, and k-means clustering \cite{hartigan1979algorithm}. In the present paper, to process incomplete data, we make use of the \emph{mean imputation approach}: for each feature, we interpolate the missing values using the average of the data available. 


After the application of the mean imputation approach, this paper proposes the modelling of the reliability of the data through a third objective of the multi-objective optimization problem. More specifically, unlike the studies in the literature, this paper not only considers the classification accuracy and solution size, but also introduces the missing rate for FS in order to enhance upon the reliability of FS. Thus, the problem is modelled as a three-objective optimization problem. Since the proposed model causes an increase in the complexity of the problem, we propose the use of non-dominated sorted genetic algorithm III (NSGA-III) \cite{deb2013evolutionary}.

The remainder of the paper is organized as follows. Section II describes the imputation method to pre-process the data. In Section III, we briefly outline the NSGA-III algorithm in the context of FS. 
Section IV describes the experimental design while Section V gives the experimental results. Section VI,  provides the conclusion of this study. 

\section{Mean Imputation Method}

\label{sec:2.3}
Missing data is a frequent problem in  machine learning. The FS methods should be correspondingly changed if the data sets have missing data. When the missing rate of the data set is less than 1\%, the influence on experimental results can be ignored. The missing rate of 1\%  $\sim$ 5\% will slightly affect the experimental results, but it can be controlled. However, if the missing rate is greater than 5\%, the results of the experiment would be affected. Therefore, for obtaining the reliable results, we need to use the effective values to estimate the missing values. Some methods have been proposed to deal with the missing data. Armina $et$ $al$. \cite{armina2017review} summarized some imputation methods for missing values.
 For example, Krause $et$ $al$. \cite{krause2018missing} designed amultiple imputation based on sophisticated imputation models. Amiri $et$ $al$. \cite{amiri2016missing} introduced a fuzzy-rough methods to handle missing data. Donder $et$ $al$. \cite{donders2006gentle} introduced some imputation methods such as single imputation and multiple imputation to get complete data sets. In this study we employ the mean imputation method in single imputation to interpolate the missing data. We chose this method since it is well-suited to handle large data sets thanks to its low computational complexity and hence modest execution time, see \cite{donders2006gentle}. The mean imputation method is divided into fixed distance imputation method and non-distance imputation method \cite{zhang2016missing}. This paper uses non-distance mean imputation method which is described as follows. 
 
 Let $\left[v_{i,j}\right]$ be the incomplete data set which is here interpreted as a matrix where some of the entries are empty, $v_{i,j}=\emptyset$ for some $i$ and $j$. Those entries that are not empty are normalised between $0$ and $1$, i.e. $v_{i,j}\in \left[0,1\right]$. The row index $i$ indicate the instance whilst the column index $j$ indicates the $j^{th}$ feature. The mean imputation method used in this study estimates the missing entries alongside the column $j$ by replacing the empty entries with $Ave_j$ calculated in the following way:
\begin{equation}
Ave_j = \frac{\sum\limits_{i= 1}^{N}  {v_{i,j}}}{N-lm_j}
\end{equation}
where $lm_j$ is the number of missing entries associated with the feature $j$, $N$ is the total number of all instances.




%

\section{Three-Objective Feature Selection Problems and NSGA-III Algorithm}

The FS problem is encoded as a multi-objective optimisation problem where its candidate solution is represented by a vector of  real numbers. Let us consider a data set with $n$ features  
\begin{equation}
\mathbf{x} = (x_{1},x_{2},\dots,x_{n})
\end{equation}
\noindent where $x_{i}\in[0,1]$. It must be remarked that the candidate solution $\mathbf{x}$ has the same structure of the row vector of the data set $\mathbf{v_i}=\left(v_{i,1},v_{i,2},\ldots,v_{i,n}\right)$. 


In order to evaluate the candidate solution $\mathbf{x}$, the objective functions are calculated in the following way. At first, the binary vector 
\begin{equation}
    \mathbf{z}=\left(z_1,z_2,\ldots,z_n\right)
\end{equation}
\noindent is generated by means of the equation 

\begin{equation}\label{eq:theta}
\scalebox{1}{$
	z_{i} = \left\{ {\begin{array}{*{20}{c}}
		{1, \begin{array}{*{20}{c}}
			{}&{\begin{array}{*{20}{c}}
				{x_{i} \geq \theta}
				\end{array}}
			\end{array}}\\
		{0,\begin{array}{*{20}{c}}
			{}&{x_{i}<\theta}
			\end{array}}
		\end{array}} \right.
	$}
\end{equation}
\noindent where $\theta$ is a threshold value  that determines whether or not a feature is selected. 

When the vector element (design variable) $x_{i}$ is greater than $\theta$ then $z_{i}$ is set equal to 1. The assignment $z_{i}=1$  denotes that the $i^{th}$ feature is selected. Conversely, if the vector $x_{i}$ is smaller than $\theta$ then $z_{i}$ is set equal to 0. The assignment $z_{i}=0$  represents that the $i^{th}$ feature is not selected. In other words, the candidate solution $\mathbf{x}$ can be interpreted as a vector whose elements represent the probability of a feature to be selected (or discarded). 


Then, with the generated $\mathbf{z}$ that represents the data set after some features (columns) have been removed, three objectives are calculated. These three objectives aim to assess: 1) classification accuracy; 2) solution size; 3) missing rate.  


For the first objective, i.e. errors of classification accuracy, we used the K-NN classifier, with $k$ set to $5$, \textcolor{black}{and we implemented $l$-fold cross-validation method ($l$=10).} The formula to calculate the classification accuracy is given as follows:
\begin{equation}
\textcolor{black}{A_{cor}= \left(\frac{{1}}{l} \sum\limits_{i = 1}^{l}  \frac{{N_{Cor}}}{{N_{All}}}\right) \times 100\%}
\end{equation}
where $N_{Cor}$ denotes the amount of test samples that are correctly predicted, $N_{All}$ denotes the number of all test samples. However, in this paper, as we use the non-dominant relationship for comparison, we introduced the \textcolor{black}{classification error rate} to evaluate the performance. The formula to calculate the  \textcolor{black}{classification error rate} and thus the first objective $f_1$ is given by:
 
\begin{equation}
f_{1}(\mathbf{x}) = 1 - A_{cor}
\end{equation}

The solution size is another objective that can be formulated as follows:
\begin{equation}
f_{2}(\mathbf{x}) = \sum\limits_{i= 1}^{n}  {z_{i}}
\end{equation}
In other words, one of the criteria is to remove as many features as possible that is to have as many zeros as possible within the vector $\mathbf{z}$. 

Classification error rate and solution size are two objectives commonly used in traditional multi-objective FS problems. 
Besides these two objectives, we introduce in this study the missing rate as the third objective. Thus, the FS problem is extended into a three-objective FS problem. The purpose of adding the third objective is to consider the reliability of the selected features. The missing rate refers to the percentage of missing data in the selected feature set with respect to  the missing data in the original data set. \textcolor{black}{At first. we store the serial number of the selected feature into the vector $y$. The number of missing data in the selected feature set is indicated with $lm$, and it is calculated as:}

\begin{equation}
lm = \sum\limits_{j= 1}^{f_{2}(\mathbf{x})}  {lm_{y_j} }
\end{equation}
\textcolor{black}{where $lm_{y_j}$ shows how many missing values in the ${y_j}^{th}$ feature.} Next, the following formula shows how we calculate the number of missing values in the original data set $la$:

\begin{equation}\label{eq:la}
la = \sum\limits_{j= 1}^{n}  {lm_{j}}
\end{equation}
After having obtained $lm$ and $la$, the missing rate can be calculated as follows:	

\begin{equation}
f_{3}(\mathbf{x}) = \frac{{l_{m}}}{l_{a}}\times 100\%
\end{equation}

\subsection{NSGA-III algorithm}
NSGA-III \cite{deb2013evolutionary} is a popular algorithm for multi-objective optimization. It's main feature is the so-called reference point-based selection method.

\begin{algorithm}
	\caption{NSGA-III algorithm }\label{alg:NSGAIII}
	\LinesNumbered 
	\KwIn{reference points $R$, parent population $P_t$}
	\KwOut{$P_{t+1}$}
	{
		{Initialize $S_t$ = 0, $i$ = 1}
		
		$Q_t$=Recombination+Mutation($P_t$);
		
		$R_t$=$P_t \cup Q_t$
		
		$(F_1,F_2,...)$=Fast-nondominated-sort($R_t$)
		
		\Repeat{$|S_t| \geq N$}{$S_t=S_t \cup F_i$ and $i=i+1$}
		
		Last front to be included: $F_l=F_i$
		
		\If{$|S_t| = N$}{
			$P_{t+1}=S_t$, break\
		}
		\Else{
			$P_{t+1}=S_t= \cup_{j=1}^{l-1}F_j$
			
			Select the point from $F_l:K=N-|P_{t+1}|$

			Based on the selection of reference points: $P_{t+1}:Selection(K, S_t, \rho_j, R, F_l)$

		}

	}
	
\end{algorithm}

Briefly, the basic idea of the NSGA-III is described as follows: firstly, it constructs a set of reference points, and randomly generates an initialization population $P_t$ of $N$ individuals, then uses binary crossover and polynomial mutations to generate new populations $Q_t$, and combines the $P_t$ and the $Q_t$ for fast non-dominated sorting. After that, $N$ individuals are chosen to enter the offspring population through the non-dominated rank, and the reference point mechanism is used for selection in the case when the selection cannot be made through the non-dominated rank \cite{deb2013evolutionary}. The structure of NSGA-III is outlined in Algorithm \ref{alg:NSGAIII}.

\subsection{Fast non-donminated sorting}

\begin{algorithm}
	\caption{Fast-nondominated-sort($R_t$)}\label{alg:nondom-sort}
	\LinesNumbered 
	\KwIn{population $P$}
	\KwOut{$F_{i}$}
	{
		
		{Initialize $S_p= \phi$, $np$ = 0}

		\For{each $p \in P$}{
			
			\For{each $q \in P$}{
				Compare p with q:	 
				\If{$p$ dominates $q$}{
					$S_p=S_p \cup \{q\}$
				}
				\Else{
					$ np=np+1$
				}

			} 
			\If{$np=0$}{
				
				$F_1=F_1 \cup \{p\}$
					
			$p_{rank}=1$
				
			}
		}
		$i=1$

		\While{$F_i \neq \phi$}{
			$Q=\phi$
			
			\ForEach{$p \in F_I$} {
				\ForEach{$q \in S_p$} {
					$n_q$=$n_q - 1$
					
					\If{$n_q=0$}{
						$q_{rank}$=$i$+1
						
						$Q=Q \cup \{q\}$
					}
					
				}
				
			}
			$i$=$i$+1
			
			$F_i$=$Q$
			
		}

	}
	
\end{algorithm}

The most important part of the fast non-dominated sorting is the non-dominated relationship \cite{deb2002fast}. When comparing non-dominated relationships, two parameters need to be calculated. The first one is $n_p$ to count the amount of individuals which dominate $p$, and the second is $S_p$ to store the individuals which  dominated by $p$. The process of fast non-donminated sorting is described as follows:

First, we initialize $S_p$ and $n_p$ by setting $S_p$ to null set and $n_p$ to 0. Then, the algorithm traverses each individual in the population $P$, and compare this individual to all the remaining individuals in $P$. For example, when it traverses to $p$, it compares $p$ with $q$. If $p$ dominates $q$, $q$ is added to $S_p$. If $q$ dominates $p$, $n_p$ is increased by 1. After traversing all the individuals, if $n_p$ = 0, $p$ is put into $F_1$. Next, it initializes the rank number $i$ to 1. Finally, for each individual $p$ in $F_1$, all individuals $q$ in $S_p$ are traversed. Whenever traversing to $q$, $n_q$ is reduced by 1. When $n_q = 0$, $q$ will be put into $F_2$ and rank of $q$ is set to 2. Next, the rank is increased by 1 and the algorithm enters the loop until $F_i = \phi$.

\subsection{Selection method based on reference points}

\begin{algorithm}[h]
	\caption{Selection($K$, $S_t$, $\rho_j$, $R$, $F_l$)}\label{alg:selection}
	\LinesNumbered 
	\KwIn{$K$, $S_t$, $\rho_j$ (the number of individuals associated with reference point $j$ in $F_1$ to $F_{l-1}$), $R$, $F_l$}
	\KwOut{$P_{t+1}$}
	{
		
		\For{each $s \in S_t$}{
			\For{each $r \in R$}{
				Compute $V(s,r) = s - r^\mathrm{T}s /|| r||$

			}
			
			Obtain the closest reference point $\pi(s)$ with individual $s$ :  $\pi(s) =r : argmin_{r \in R} V(s,r)$
			
			Obtain the  distance between $s$ and  $\pi(s)$: $d(s) = V(s,\pi(s) )$

		}
		
		$\rho_j=\sum_{s\in F_c}(\pi(s)=j)  $ where $c=1,...,l-1$

		$k$=1

		\While{$k < K$}{
			$J_{min}$=$\{j:argmin_{j \in R} \rho_j\}$
			
			$j_r$=random($J_{min}$)
			
			$N_j$ = $\{s: \pi (s) = j_r, s \in F_l\}$

			\If{$N_j \neq 0$}{
				
				\If{$\rho_j=0$}{
					
					$P_{t+1}$  = $P_{t+1}$   $\cup \{s\}$  ($s:argmin_{s \in N_j}d(s)$)
				}
				\Else{$P_{t+1}$  = $P_{t+1}$   $\cup random(N_j)$}
				
				$\rho_j = \rho_j + 1$
				
				$F_l = F_l / s$
				
				$k = k + 1$

			}
			\Else{
				$R=R	/ j_r$
			}
			
		}

	}
	
\end{algorithm}
After the fast non-donminated ranking, individuals are put into the next offspring population according to the non-donminated rank. When the offspring individuals cannot be selected with the non-donminated rank, we use the reference point selection method. First, we construct the reference points by using the methods devised by  Das and Denniss  \cite{das1998normal}. This method generates $C_{P+M-1}^{P}$ isometric reference points on an equilateral triangle whose apexes are (1,0,0), (0,1,0) and (0,0,1), and this coordinate axis is based on the ideal point $p^{ideal}_j$ as the origin. $M$ is the objective dimension. Each objective is divided into $P$ parts. Fig.1 depicts an example of a reference point set with three objectives and each objective is separated into Four parts.

\begin{figure}[h]
	\includegraphics[width=0.36\textwidth]{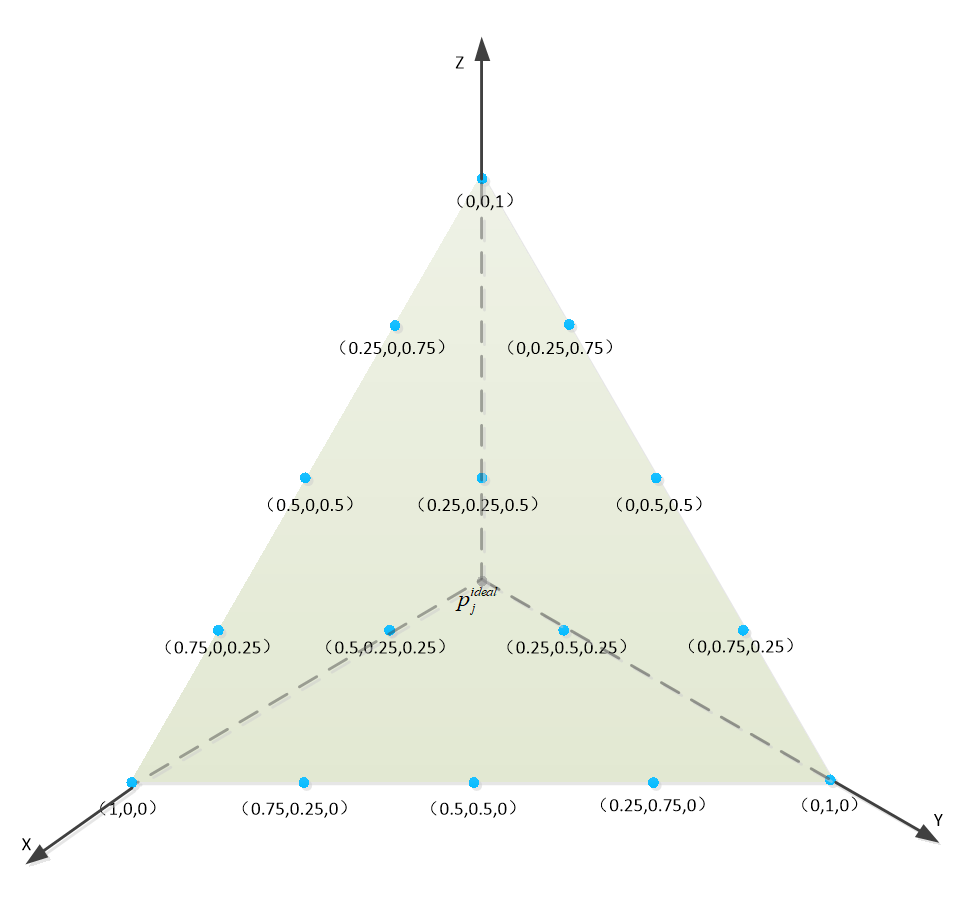}
	\centering
	\caption{ The coordinates of reference points with three objectives and each objective divided into four parts.}
	\label{fig:1}	
\end{figure}

The distance between each reference point is $\frac{{1}}{P}$, and the coordinate $r_j$ calculation formula of each reference point is given as follows.

\begin{equation}
r_{j}= (r_{1},r_{2},...,r_{j}) , j=1,2,...,M
\end{equation}

\begin{equation}
r_j \in \left\{ 0,1/P,...,P/P \right\}  ,\sum\limits_{j= 1}^{M}{r_j=1}
\end{equation}

Then each individual needs to link to a reference point. Firstly, we compute the reference line that is between the reference point and the origin $p^{ideal}_j$. Then, we calculate the vertical distance $V$ from each individual in the population $S_t$ to each reference line, and the formula of vertical distance $V$  is given as follows:

\begin{equation}
V(s,r) = s - r^\mathrm{T}s / ||r||
\end{equation}
Finally, we associate the individual with the reference point corresponding to the closest reference line.

When we add the individuals to the next generation $P_{t+1}$  from $F_l$. Firstly, we randomly select a reference point with the least individual association in $F_1$ to $F_{l-1}$, and then obtain the $N_j$, where $N_j$ represents the amount of individuals associating with the $j_r$ in the current frontier $F_l$. Next, if $N_j=0$ (it means that no individual in $F_l$  associates with the $j_r$), and then a $j_r$ is replaced. But if $N_j\neq0$ (it means that there are individuals in $F_l$ associating with the $j_r$), and next if $\rho_j=0$ (it means that no individual associates with $j$ between $F_1$ and $F_{l-1}$), the individual with the shortest distance is selected. Otherwise, an individual is randomly selected.

Algorithm \ref{alg:selection} shows the pseudo-code of NSGA-III selection.

\section{Experimental Setup}

Table \ref{tab:1} displays the six incomplete data sets from the University of California Irvine (UCI) Machine Learning Repositor. 

With reference to Table \ref{tab:1}, DS$i$ represents the $i^{th}$ data set, DN represents the names of data sets.  NoI indicates the number of individuals in the data sets, Dim denotes the data set dimension, and AoC denotes the amount of classes, MR denotes the percentage of missing values in the data sets. It can be observed that the highest missing rate is 24.9\%.

\begin{table}
	
	\newcommand{\tabincell}[2]{\begin{tabular}{@{}#1@{}}#2\end{tabular}}
	
	\centering
	\caption{Information of Datasets}
	\label{tab:1}       
	\doublerulesep=3.0pt
	\begin{lrbox}{\tablebox}
		\begin{tabular}{ c c c c c c}
			\hline\noalign{\smallskip}
			NO.&DN&NoI&Dim&AoC&MR\\
			\hline\noalign{\smallskip}
			DS1&	processed.va.data&	200&	14&	5 &24.9\%\\
			DS2&	Heart-h&	294&	14&	2&19.0\%	\\
			DS3&	Hepatitis&	155&	20&	2&5.4\%\\
			DS4&	Tumor.data&	339&	18&	21&3.7\%\\	
			
			DS5&	\tabincell{c}{processed.\\switzerland.data}&	123&	14&	4&15.8\%\\

			DS6&	arrhythmia.data&	452&	279&	16&6.0\%\\

			\hline\noalign{\smallskip}
		\end{tabular}
	\end{lrbox}
	\scalebox{1}{\usebox{\tablebox}} 
\end{table}

\subsection{Preparation work}
First, we use the mean imputation method to fill in the missing values in the data sets. Then, we separate each data set into two parts, i.e., the training set and test set. 70\% of the examples of the initial data set are chosen as the training set at random, and the remaining examples are used as test set. K-NN method is utilized for evaluating the fitness value of the feature subset and the 10-fold cross-validation is utilized for measuring classification accuracy. 


\subsection{Benchmark algorithms and parameter settings}

To verify the effectiveness of NSGA-III on three objectives FS problems, four algorithms of NSGA-II \cite{deb2002fast}, SPEA-II \cite{shi2015multi}, IBEA \cite{li2019ibea} and KnEA \cite{zhang2014knee} are used for comparison. These comparison algorithms are run within the PLATEMO \cite{tian2017platemo} platform. The specification of the platform is fundamental in accordance with the study reported in \cite{bib:Rostami2020}.  Each algorithm has run 30 times on the six data sets. Each run has been stopped when the computational budget on the number of fitness evaluation ($NFE$) was reached.

With reference to \cite{deb2013evolutionary}, the parameters of NSGA-III are set in the following way: $NFE$ = 100000, $\theta$ = 0.6 see eq. (\ref{eq:theta}), number of objectives $M$ = 3 , population size $PS$ = 100, the upper bound of individuals $up$ = 1, and the lower bound $ub$ = 0.

\subsection{Performance metrics}
In order to evaluate the performance of the NSGA-III algorithm on the multi-objective FS problems, we introduce two indicators. The first is the inverted generational distance (IGD) \cite{sun2018igd}. The descriptions of the IGD is given as follows:

\begin{equation}
IGD(D,Z)=\frac{1}{|Z|}\sum\limits_{i= 1}^{|z|}{min_{j=1 to |D|}ed(z_i,d_i)}
\end{equation}
where $D$ denotes the non-dominated solution set, and $Z$ is the objective solution set. $ed(z_i,d_i)$ represents the minimum Euclidean distance from the individual in $Z$ to population $D$. The smaller the value of IGD, the better the distribution and convergence quality of the solutions.

The second is hyper-volume metric (HV) \cite{DBLP:journals/swevo/RostamiN17}. HV is a kind of quality judgment of the test algorithm by comprehensively evaluating the convergence, extensiveness and distribution of the solution set of the multi-objective optimization algorithm, see \cite{DBLP:journals/icae/RostamiNE17}.

\section{Experimental Results}
	
	\begin{table*}
		\centering
		\caption{MEAN VALUES AND STANDARD DEVIATIONS OF IGD VALUES OBTAINED BY THE FIVE ALGORITHMS ON THE TRAINING SETS}
		\label{tab:3}
		\doublerulesep=0.0pt
		\resizebox{\textwidth}{25mm}{
			\begin{lrbox}{\tablebox}
				\begin{tabular}{  c  c  c  c c c c  c c  c c c }
					\hline\noalign{\smallskip}			
					\multicolumn{2}{c}{\multirow{2}{*}{Datasets}}&\multicolumn{2}{c}{NSGA-II}&\multicolumn{2}{c}{SPEA-II}&\multicolumn{2}{c}{IBEA}&\multicolumn{2}{c}{KnEA}&\multicolumn{2}{c}{NSGA-III}\\
					\cline{3-12}\noalign{\smallskip}
					&&MV &SD &MV &SD &MV &SD &MV &SD &MV &SD\\			
					\hline
					\hline\noalign{\smallskip}
					\multirow{2}{*}{DS1}&IGD& \textbf{1.87E-02} & 8.02E-03	&3.60E-02 &7.86E-02& 2.87E-01 & 2.64E-01 &3.13E-02  &0.0497 &9.04E-02&	8.97E-02 	\\
					
					&T-sig&-&&-&&+&&-&& &\\
					\multirow{2}{*}{DS2} &IGD& 0.105 & 9.50E-02 & 5.04E-02 &5.32E-02 & 4.51E-01 &2.57E-01  &\textbf{4.90E-02}  &4.87E-02 &4.97E-02	&	4.75E-02	\\
					
					&T-sig&+&&=&&+&&=&& &\\
					\multirow{2}{*}{DS3}&IGD & 2.12E-01 & 9.50E-02 & 7.90E-02 &4.33E-02 &1.55E-01  &5.91E-02  & \textbf{5.28E-02} & 2.31E-02&1.60E-01	&	6.80E-02	\\
					&T-sig&+&&-&&=&&-&& &\\
					
					\multirow{2}{*}{DS4}&IGD & 8.22E-02 & 4.96E-02& 7.07E-02 &1.74E-02 &5.49E-02  &2.82E-02  & \textbf{1.66E-02} &9.85E-03 	&4.71E-02  &	3.69E-02	\\
					&T-sig&+&&+&&=&&-&& &\\
					
					\multirow{2}{*}{DS5} &IGD& 6.41E-04 & 3.56E-04	&7.58E-02  &1.49E-01 &4.31E-03  &1.20E-03  & 5.51E-02 & 1.18E-01	&\textbf{3.20E-04}	&	2.83E-04	\\
					&T-sig&+&&+&&+&&+&& &\\
					
					\multirow{2}{*}{DS6}&IGD& 4.87E+00 & 3.83E+00	&4.11E+00  &2.34E+00 &2.60E+00  &1.62E+00  &2.33E+00  &2.10E+00 &\textbf{1.57E+00}	&	2.28E+00 	\\
					&T-sig&+&&+&&+&&=&& &\\

					\hline\noalign{\smallskip}			
				\end{tabular}
			\end{lrbox}
			
			\scalebox{1}{\usebox{\tablebox}} 
		}
	\end{table*}
	\begin{table*}
		\centering
		\caption{MEAN VALUES AND STANDARD DEVIATION OF IGD VALUES OBTAINED BY THE FIVE ALGORITHMS ON THE TEST SETS}
		\label{tab:4}       
		\doublerulesep=0.0pt
		\resizebox{\textwidth}{25mm}{
			\begin{lrbox}{\tablebox}
				\begin{tabular}{  c  c  c  c c c c  c c  c c c }
					\hline\noalign{\smallskip}			
					\multicolumn{2}{c}{\multirow{2}{*}{Datasets}}&\multicolumn{2}{c}{NSGA-II}&\multicolumn{2}{c}{SPEA-II}&\multicolumn{2}{c}{IBEA}&\multicolumn{2}{c}{KnEA}&\multicolumn{2}{c}{NSGA-III}\\
					\cline{3-12}\noalign{\smallskip}
					&&MV &SD &MV &SD &MV &SD &MV &SD &MV &SD\\			
					\hline
					\hline\noalign{\smallskip}
					\multirow{2}{*}{DS1}&IGD	&   1.85E-01	 &1.53E-01 &3.05E-01  &1.94E-01  &1.21E-01  &1.35E-01  &4.44E-01  &3.31E-01 &\textbf{1.20E-01}	&   1.09E-01 	\\
					
					&T-sig&+&&+&&=&&+&& &\\
					\multirow{2}{*}{DS2}&IGD   & 3.61E-01    &1.96E-01	
					&7.09E-01  &1.18E-01  &1.10E+00  &3.13E-01  &	\textbf{3.38E-01}  & 1.58E-01 &3.83E-01	&	2.19E-01 	\\
					
					&T-sig&=&&+&&+&&=&& &\\
					\multirow{2}{*}{DS3}&IGD	& \textbf{2.91E-01} &  2.17E-01
					&4.49E-01  &4.49E-01  &3.47E-01  &1.66E-01  &	4.67E-01  &  3.00E-01&5.77E-01	&	2.02E-01	\\
					&T-sig&-&&=&&+&&=&& &\\
					
					\multirow{2}{*}{DS4}&IGD	& 5.24E-01	& 2.65E-01	
					&\textbf{3.09E-01}  &2.12E-01  &4.74E-01  &2.55E-01  & 3.73E-01 & 2.58E-01 	&3.65E-01	&	2.00E-01\\
					&T-sig&+&&=&&=&&=&& &\\
					
					\multirow{2}{*}{DS5}&IGD	& 1.31E+00	 & 	5.19E-01
					&6.51E-01  &5.71E-01  &\textbf{2.90E-01}  &2.50E-01  &3.91E-01  &3.35E-01  &7.14E-01	&   4.60E-01	\\
					&T-sig&+&&-&&-&&-&& &\\
					
					\multirow{2}{*}{DS6}&IGD	& 8.86E+00	 & 2.29E+00	
					&6.63E+00  &3.96E+00  &4.96E+00  &1.17E+00  &3.67E+00  & 2.03E+00 	&\textbf{3.47E+00}	&	1.72E+00\\
					&T-sig&+&&+&&+&&=&& &\\

					\hline\noalign{\smallskip}			
				\end{tabular}
			\end{lrbox}
			\scalebox{1}{\usebox{\tablebox}} 
		}
	\end{table*}

The experimental results on the six data sets are listed in \textcolor{black}{Table} \ref{tab:3}, \ref{tab:4}, \ref{tab:5}, \ref{tab:6}. In these \textcolor{black}{Tables}, MV indicates the mean value while SD indicates the standard deviation. The best mean values are highlighted in bold. T-sig indicates the statistical significance of the results according to the T-test with confidence level $95\%$.  The ``+" denotes that NSGA-III is significantly better than the comparison approach, the ``-" denotes that the comparison approach is significantly better than NSGA-III, and ``=" denotes that NSGA-III and comparison approach have similar results.

Numerical results on the training sets in \textcolor{black}{Table} \ref{tab:3} show that the mean values obtained by NSGA-III on the six data sets are smaller, and the corresponding standard deviations are smaller too.  Through comparing NSGA-III with NSGA-II, SPEA-II, IBEA and KnEA, it is found that NSGA-III is significantly better than NSGA-II and  it outperforms NSGA-II on five data sets with significant difference. NSGA-III performs similar to NSGA-II on one training set. NSGA-III is superior to SPEA-II on three training sets with significant difference. The results of NSGA-III are similar to SPEA-II on two training sets. When comparing NSGA-III with IBEA, it is found that the IGD values obtained by NSGA-III on the four training sets are smaller than IBEA with significant difference. The performance of NSGA-III on two training sets is similar to that of IBEA. The IGD value of NSGA-III is smaller than KnEA on one data set with significant difference, and is similar to KnEA on two training sets. Through the above analysis, we can get that NSGA-III is better than NSGA-II, SPEA-II and IBEA when using IGD index, and it has  similar performance to that of KnEA. By analyzing \textcolor{black}{Table} \ref{tab:4}, we can reach the same conclusion on the test sets: NSGA-III is still superior to NSGA-II, SPEA-II and IBEA, and it has a similar performance to that of KnEA.


\begin{table*}
	\centering
	\caption{MEAN VALUES AND STANDARD DEVIATIONS OF HV VALUES OBTAINED BY THE FIVE ALGORITHMS ON THE TRAINING SETS}
	\label{tab:5}       
	\doublerulesep=0.0pt
	
	\resizebox{\textwidth}{25mm}{
		\begin{lrbox}{\tablebox}
			
			\begin{tabular}{  c  c  c  c c c c  c c  c c c }
				\hline\noalign{\smallskip}			
				\multicolumn{2}{c}{\multirow{2}{*}{Datasets}}&\multicolumn{2}{c}{NSGA-II}&\multicolumn{2}{c}{SPEA-II}&\multicolumn{2}{c}{IBEA}&\multicolumn{2}{c}{KnEA}&\multicolumn{2}{c}{NSGA-III}\\
				\cline{3-12}\noalign{\smallskip}
				&&MV &SD &MV &SD &MV &SD &MV &SD &MV &SD\\			
				\hline
				\hline\noalign{\smallskip}
				\multirow{2}{*}{DS1}&HV & 5.38E-02	 & 2.77E-03	&4.87E-02  &3.25E-03  &4.36E-02  &5.12E-02  &5.35E-02  &2.68E-03  &\textbf{1.78E-01}	&	8.52E-03\\
				
				&T-sig&+&&+&&+&&+&& &\\
				\multirow{2}{*}{DS2}&HV& 6.29E-02 & 4.71E-03	&1.38E-01  &6.64E-03  &1.18E-01  &1.27E-01  &1.15E-01  &6.13E-03 &\textbf{1.51E-01}	&	7.72E-03 	\\
				
				&T-sig&+&&+&&+&&+&& &\\
				\multirow{2}{*}{DS3}&HV & 2.41E-01 &  8.25E-03	&1.91E-01  &5.02E-03  &1.92E-01  &2.13E-01  &1.75E-01  & 3.47E-03&\textbf{2.64E-01}	&	6.74E-03	\\
				&T-sig&+&&+&&+&&+&& &\\
				
				\multirow{2}{*}{DS4}&HV & \textbf{1.59E+00} & 1.12E-02	 &1.92E+00  &1.54E-02  &5.01E-01  &5.16E-01  &5.18E-01 &4.82E-03&1.35E+00	&	8.39E-03 \\
				&T-sig&-&&-&&+&&+&& &\\
				
				\multirow{2}{*}{DS5}&HV&1.04E-01 &2.97E-03	&8.32E-02  &2.12E-03  & 5.32E-04 &5.57E-04  & \textbf{1.10E-01}  &4.26E-03 	&1.03E-01	&2.00E-03 	\\
				&T-sig&=&&+&&+&&-&& &\\
				
				\multirow{2}{*}{DS6}&HV & 3.92E-02 & 1.92E-02		&\textbf{1.16E-01}  &1.60E-02  &3.22E-03  &5.49E-03  & 2.73E-03 &8.94E-04 &3.27E-02	&	1.11E-02 	\\
				&T-sig&=&&-&&=&&+&& &\\
				
				\hline\noalign{\smallskip}			
			\end{tabular}
		\end{lrbox}
		\scalebox{1}{\usebox{\tablebox}} 
	}
\end{table*}

\begin{table*}
	\centering
	\caption{MEAN VALUES AND STANDARD DEVIATIONS OF HV VALUES OBTAINED BY THE FIVE ALGORITHMS ON THE TEST SETS}
	\label{tab:6}       
	
	\doublerulesep=0.0pt
	\resizebox{\textwidth}{25mm}{
		\begin{lrbox}{\tablebox}
			\begin{tabular}{  c  c  c  c c c c  c c  c c c }
				\hline\noalign{\smallskip}			
				\multicolumn{2}{c}{\multirow{2}{*}{Datasets}}&\multicolumn{2}{c}{NSGA-II}&\multicolumn{2}{c}{SPEA-II}&\multicolumn{2}{c}{IBEA}&\multicolumn{2}{c}{KnEA}&\multicolumn{2}{c}{NSGA-III}\\
				\cline{3-12}\noalign{\smallskip}
				&&MV &SD &MV &SD &MV &SD &MV &SD &MV &SD\\			
				\hline
				\hline\noalign{\smallskip}
				
				\multirow{2}{*}{DS1} &HV & 5.09E-04	&3.10E-04	
				&3.30E-04  &3.20E-04  &0.00E+00  &0.00E+00  &5.06E-04  & 	2.97E-04 & \textbf{2.24E-02}	&	4.91E-02 	\\
				
				&T-sig&+&&+&&+&&+&& &\\
				\multirow{2}{*}{DS2}&HV & 4.06E-02 &4.80E-03	&1.67E-02  &1.78E-03  &\textbf{2.80E-01}  &2.54E-02  &3.80E-02  & 4.54E-03 	& 1.09E-01	&   1.06E-02   \\
				&T-sig&+&&+&&-&&+&& &\\
				
				\multirow{2}{*}{DS3}&HV &1.32E-03 &1.20E-03&5.04E-03  &1.95E-03  &3.85E-03  & 2.07E-03 &5.23E-03  & 	2.04E-03 & \textbf{1.23E-02}  &	3.33E-03 	\\
				&T-sig&+&&+&&+&&+&& &\\
				
				\multirow{2}{*}{DS4}&HV& \textbf{ 4.42E-01} &2.91E-02		
				&3.26E-01  &2.01E-02  &3.84E-01  &2.59E-02  &3.83E-01  &2.56E-02  &3.30E-01	&	1.73E-02 	\\
				&T-sig&-&&=&&-&&-&&\\
				
				\multirow{2}{*}{DS5}&HV& 6.19E-02 &2.51E-03		&\textbf{8.78E-02}  &3.41E-03  &5.28E-04  &2.22E-05  & 6.16E-02 &3.53E-03 & 6.46E-02	&	1.90E-03  	\\
				&T-sig&+&&-&&+&&=&& &\\
				
				\multirow{2}{*}{DS6}&HV& 	7.17E-02 & 1.74E-02	
				&7.19E-05  &2.63E-04  &	3.13E-02  &1.08E-02  & 1.17E-02 &6.81E-03 & \textbf{1.46E-01}	&	1.52E-02  	\\
				&T-sig&+&&+&&+&&+&& &\\

				\hline\noalign{\smallskip}			
			\end{tabular}
		\end{lrbox}
		\scalebox{1}{\usebox{\tablebox}} 
	}
\end{table*}

Table \ref{tab:5} and \textcolor{black}{Table} \ref{tab:6} show the results in terms of HV for the training and test sets respectively. Table \ref{tab:5} shows that for three data sets,  NSGA-III displays a significantly better performance than NSGA-II and in two cases NSGA-II and NSGA-III have a similar performance.  NSGA-III performs better than SPEA-II on four training sets. NSGA-III is superior to IBEA and KnEA on five training sets. The  results in Table \ref{tab:6} clearly show that NSGA-III is superior to other algorithms in the majority of cases  in terms of HV. According to our interpretation, the reason for the good performance of NSGA-III is that it constructs a uniformly distributed reference point system, so that the selected offspring are evenly distributed in the objective space, reducing the situation of the offspring gathering together, which improves the distribution of offspring, and also increase the diversity of offspring.

\begin{figure*}
	\includegraphics[width=0.48\textwidth]{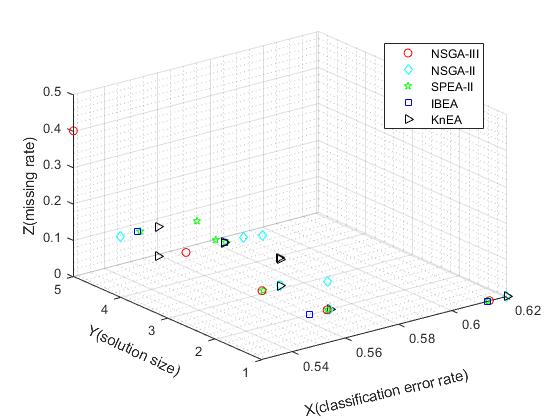}
	\includegraphics[width=0.48\textwidth]{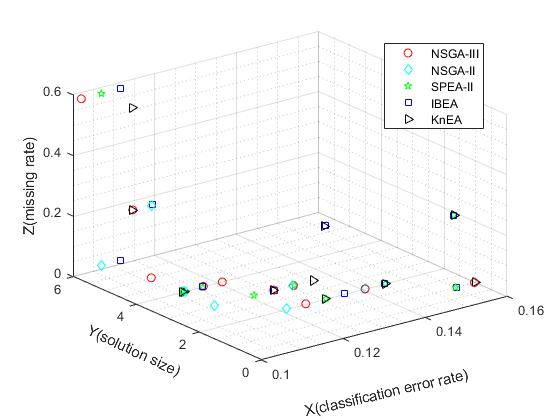}
	\centerline{~~~~ DS 1~~~~~~~~~~~~~~~~~~~~~~~~~~~~~~~~~~~~~~~~~~~~~~~~~~~~~~~~~~~~~~~~~~~~~~~~~~~~~~~~~~~~~~~~~~ DS 2}
	\includegraphics[width=0.48\textwidth]{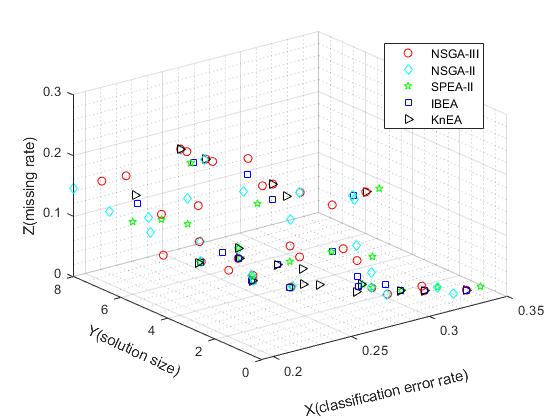}
	\includegraphics[width=0.48\textwidth]{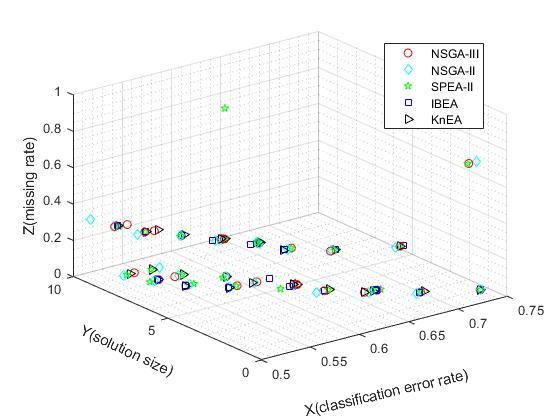}
	\centerline{~~~~ DS 3~~~~~~~~~~~~~~~~~~~~~~~~~~~~~~~~~~~~~~~~~~~~~~~~~~~~~~~~~~~~~~~~~~~~~~~~~~~~~~~~~~~~~~~~~~ DS 4}	
	\includegraphics[width=0.48\textwidth]{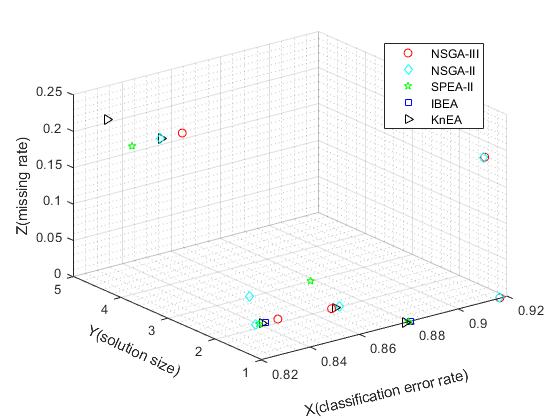}
	\includegraphics[width=0.48\textwidth]{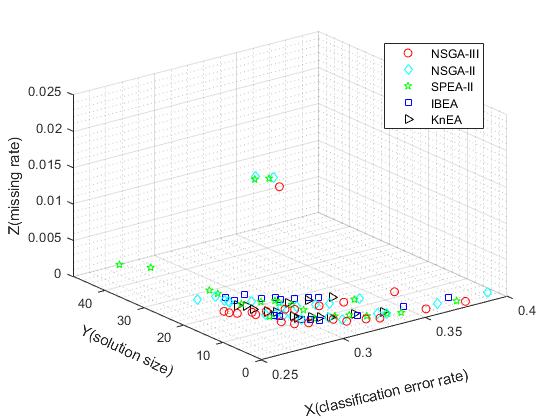}
	\centerline{~~~~ DS 5~~~~~~~~~~~~~~~~~~~~~~~~~~~~~~~~~~~~~~~~~~~~~~~~~~~~~~~~~~~~~~~~~~~~~~~~~~~~~~~~~~~~~~~~~~ DS 6}
	
	\centering
	\caption{Classification accuracy,solution size and missing rate of different algorithms on training sets (DS 1- DS 6).}
	
	\label{fig:2}
	
\end{figure*}

\begin{figure*}
	
		\includegraphics[width=0.48\textwidth]{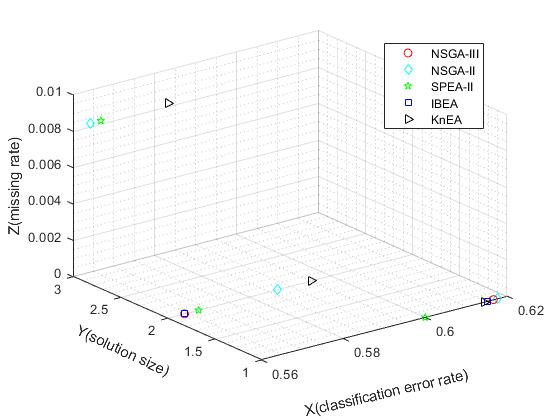}
	\includegraphics[width=0.48\textwidth]{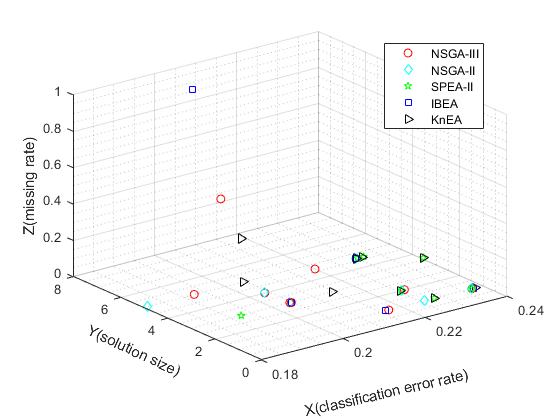}
	\centerline{~~~~ DS 1~~~~~~~~~~~~~~~~~~~~~~~~~~~~~~~~~~~~~~~~~~~~~~~~~~~~~~~~~~~~~~~~~~~~~~~~~~~~~~~~~~~~~~~~~~ DS 2}
	\includegraphics[width=0.48\textwidth]{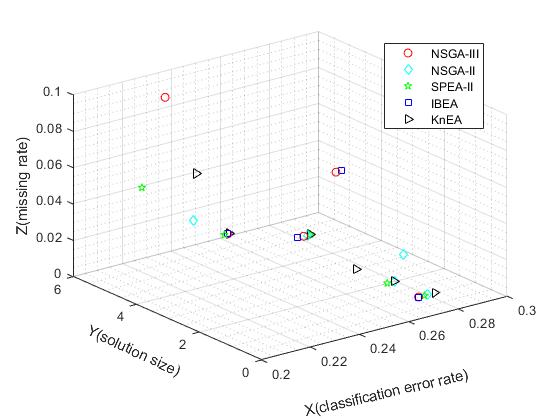}
	\includegraphics[width=0.48\textwidth]{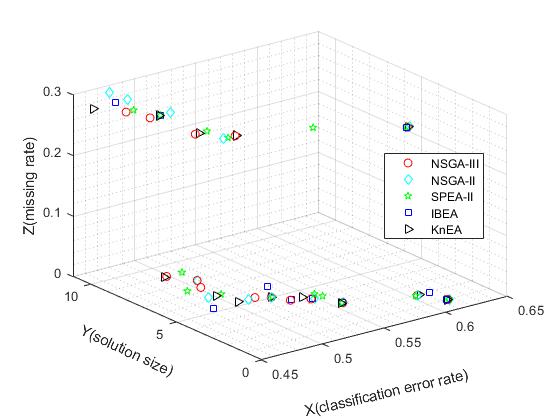}
	\centerline{~~~~ DS 3~~~~~~~~~~~~~~~~~~~~~~~~~~~~~~~~~~~~~~~~~~~~~~~~~~~~~~~~~~~~~~~~~~~~~~~~~~~~~~~~~~~~~~~~~~ DS 4}
	\includegraphics[width=0.48\textwidth]{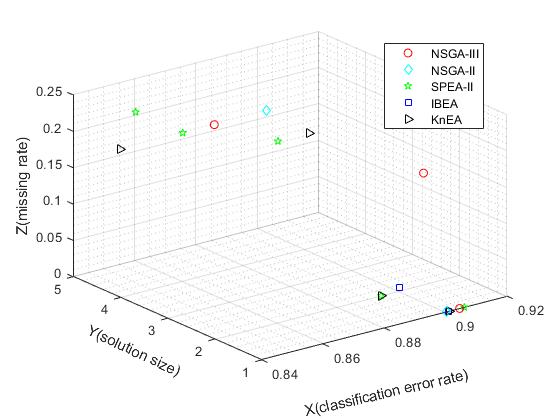}
	\includegraphics[width=0.48\textwidth]{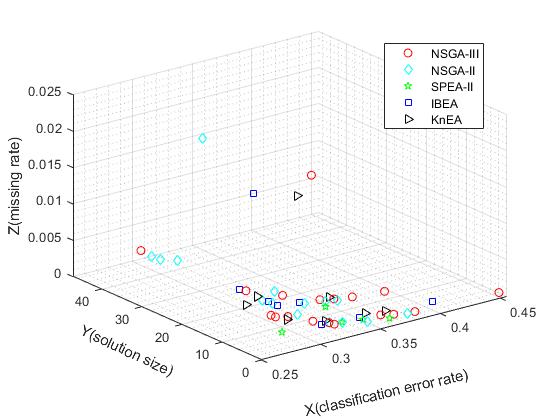}
	\centerline{~~~~ DS 5~~~~~~~~~~~~~~~~~~~~~~~~~~~~~~~~~~~~~~~~~~~~~~~~~~~~~~~~~~~~~~~~~~~~~~~~~~~~~~~~~~~~~~~~~ DS 6}
	
	\centering
	\caption{\textcolor{black}{Classification accuracy,solution size and missing rate of different algorithms on test sets (DS 1- DS 6).}}
	
	\label{fig:2a}
	
\end{figure*}

On all data sets, each algorithm has obtained a set of solutions with a non-dominated rank of 1. In order to depict the advantages of NSGA-III with respect to its competitors, Fig.\ref{fig:2} and Fig.\ref{fig:2a} show the Pareto fronts of NSGA-III and the other four algorithms considered in this study on the six data sets used in the experiments. 
Fig.\ref{fig:2} shows the results on the training sets while Fig.\ref{fig:2a} presents the results on test sets. In each subfigure,  the x-axis represents the classification error rate $f_1$, the y-axis represents the solution sizes $f_2$, and the z-axis represents the missing rate $f_3$.

Results in Fig.\ref{fig:2} and Fig.\ref{fig:2a} show that the distribution of the solutions in the non-dominated sets detected by NSGA-III are more uniform than those detected the other algorithms. Furthermore, NSGA-III is better than other algorithms in terms of classification accuracy, solution size and missing rate on both training  and test sets. On the basis of the results we obtained, we conclude that NSGA-III has better performance for FS problems with three objectives. According to our interpretation, the reason why NSGA-III is superior to other algorithms is that it uses the method of reference point selection, which associates individuals with each reference point, and effectively selects individuals with less correlation with reference points. This mechanism enables distribution uniform of the populations, which then promotes the offspring population to produce more diverse individuals in the following generations.

\section{Conclusion}

This paper proposes a novel interpretation of FS problem in data science with a specific reference to data sets with missing data. Unlike classical studies in the literature that use accuracy and size of the solutions as quality metrics, we propose the simultaneous inclusion of a third metric, that is the missing rate. This modelling poses a three-objective optimization problem that is addressed by means of an ad-hoc implementation of NSGA-III.  

In order to demonstrate the effectiveness of the proposed approach, we tested NSGA-III on six incomplete data sets from the the UCI machine learning repository and compared them against four popular algorithms for multi-objective optimization. Numerical results show the overall superiority NSGA-III to the other methods considered in this study in terms of IGD and HV.

Although the performance of our NSGA-III implementation is promising, we feel that there is some margin for improvement. Future research will investigate the integration of knowledge-based features associated to the FS to the selection mechanism of NSGA-III. 

\section*{Acknowledgements}
This work was partially supported by the National Natural Science Foundation of China (61876089,  61876185, 61902281), the opening Project of Jiangsu Key Laboratory of Data Science and Smart Software(No.2019DS301), the Science and Technology Program of Ministry of Housing and Urban-Rural Development (2019-K-141), the Engineering Research Center of Digital Forensics, Ministry of Education, the Entrepreneurial team of sponge City (2017R02002), and the PAPD.

\ifCLASSOPTIONcaptionsoff
  \newpage
\fi

\bibliographystyle{IEEEtran}
\bibliography{lalbelpaper}

\end{document}